\pdfoutput=1

\documentclass[11pt]{article}

\usepackage{emnlp2021}

\usepackage{times}
\usepackage{latexsym}

\usepackage[T1]{fontenc}

\usepackage[utf8]{inputenc}
\usepackage{microtype}

\usepackage[switch]{lineno} 

\usepackage{graphicx}
\usepackage{float}
\usepackage{flushend}
\usepackage{amsmath}   
\newcommand{\nk}[1]{\textcolor{green}{Nora: #1}}

\newcommand{\eat}[1]{}
\newcommand{\red}[1]{\textcolor{red}{#1}}

\newcommand{\edited}[1]{{#1}}

\usepackage{quoting}
\newenvironment{myquote}{                   
  \parskip 0mm \begin{quoting}[vskip=0mm,leftmargin=2mm]}{
\end{quoting}}
\newenvironment{myquote2}{                   
  \parskip 1mm \begin{quoting}[vskip=0mm,leftmargin=4mm]}{
\end{quoting}}
\newenvironment{ite}{                     
     \parskip 0cm \begin{itemize} \parskip 0cm \parsep 0cm \itemsep 0cm \topsep 0cm}{
        \end{itemize}} 
\newenvironment{enu}{                   
     \parskip 0cm \begin{list}{}{\parsep 0cm \itemsep 0cm \topsep 0cm}}{
       \end{list}} 
\newenvironment{des}{                 
     \parskip 0cm \begin{list}{}{\parsep 0cm \itemsep 0cm \topsep 0cm}}{
       \end{list}} 
\newcommand{\mybullet}[1]{$\bullet$ {\bf #1}}

\title{BeliefBank: Adding Memory to a Pre-Trained Language Model for a Systematic Notion of Belief}

\author{Nora Kassner\textsuperscript{1,2},  Oyvind Tafjord\textsuperscript{1}, Hinrich Sch{\"u}tze\textsuperscript{2}, Peter Clark\textsuperscript{1} \\
\textsuperscript{1}Allen Institute for AI, Seattle, WA \\
\textsuperscript{2}Center for Information and Language Processing, 
LMU Munich, Germany \\
\texttt{kassner@cis.lmu.de} \\
\texttt{\{oyvindt,peterc\}@allenai.org} 
}

\begin{document}
\maketitle

\begin{abstract}
Although pretrained language models (PTLMs) contain significant amounts of
world knowledge, they can still produce inconsistent answers to questions
when probed, even after specialized training.  As a result, it can be hard to
identify what the model actually ``believes'' about the world, making it
susceptible to inconsistent behavior and simple errors. Our goal is to
reduce these problems.
Our approach is to embed a PTLM in a broader system that also includes an
evolving, symbolic memory of beliefs -- a BeliefBank -- 
that records but then may modify the raw PTLM answers. 
We describe two mechanisms to improve belief consistency in the overall
system. First, a reasoning component -- a weighted MaxSAT solver -- revises
beliefs that significantly clash with others. Second, a feedback
component issues future queries to the PTLM using known beliefs as context.
We show that, in a controlled experimental setting, these two mechanisms
result in more consistent beliefs in the overall system,
improving both the accuracy and consistency of its answers over time.
This is significant as it is a first step towards PTLM-based architectures
with a systematic notion of belief, enabling them to construct a more coherent
picture of the world, and improve over time without model retraining.
\end{abstract}

\section{Introduction}

\begin{figure}[t]
\centering
     \includegraphics[width=1\columnwidth]{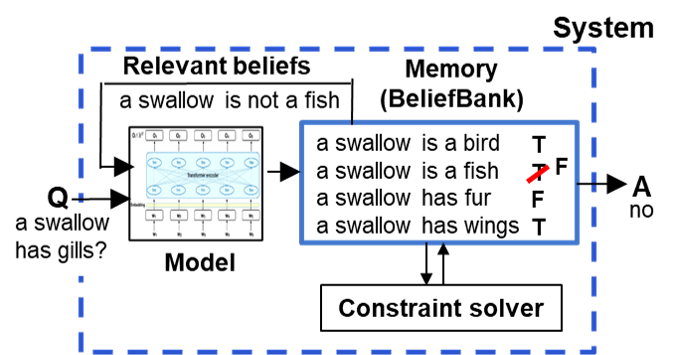}	   
      \vspace{-3mm}
\caption{ 
The proposed architecture. The model's raw answers are stored in a
persistent, symbolic memory (BeliefBank), and two mechanisms attempt to improve them:
(a) A {\bf constraint solver} flips beliefs
(e.g., the belief that ``a swallow
is a fish'')
that clash
significantly with others.
(b) A {\bf feedback} mechanism poses new questions using
existing, relevant beliefs (e.g., ``a swallow is not a fish'')
as the query context. We find that both consistency and accuracy
of the overall system improve.
Example: The model $M$ shown in the figure incorrectly answers
``yes'', when asked ``a swallow has gills?''.
But (as shown above) if reminded of its previous answer ``a swallow is not a fish'',  
$M$ correctly answers "no". 
\label{architecture}}

 \vspace{-6mm}
\end{figure}

Intelligent agents are typically considered to have beliefs about the
world -- propositions that they take as true \cite{sep-formal-belief}.
In general, a system can be said to (appear to) believe a proposition \textit{p}, e.g., ``eagles are birds'',
if it 
produces answers consistent with \textit{p} (and its other beliefs). 
Pragmatically, we expect the system to (a) give a consistent answer to
different paraphrases of the question "\textit{p}?" ("Are eagles birds?", "Is an eagle a type of bird?", ...),
and (b) give correct answers about implications of \textit{p} ("Eagles lay eggs", "Eagles have feathers", ...),
conditional on its other knowledge and reasoning abilities.

Maintaining a consistent
set of beliefs (a ``belief system'') is a key facet of intelligence, as it
can help debug errors and encourage rational behavior.
However, although PTLMs contain substantial world knowledge \cite{Petroni2019LanguageMA},
their answers to probing questions can be inconsistent \cite{Elazar2021MeasuringAI, 
Kassner2020NegatedAM},
even after specialized training to reduce inconsistency \cite{Ribeiro2019AreRR,Li2019ALF}.
As a result, it is sometimes hard to pin down what a PTLM actually ``believes'',
making them susceptible to inconsistent and/or irrational behavior.
Our goal is a first step to avoid these problems by embedding a PTLM in a broader system
with a clearer notion of belief (see Figure~\ref{architecture}). 

Prior work in AI, including in formal logic \cite{Genesereth1987LogicalFO}, belief maintenance \cite{de1986assumption,dechter1988belief},
and uncertainty \cite{Pearl1986FusionPA},
offers models for how beliefs can be managed. Most importantly, it posits that creating
a coherent 
set of beliefs -- a kind of ``mental model'' of the world \cite{JohnsonLaird1983MentalM} - 
is a constructive process requiring explicit representation of beliefs,
and inference 
about their dependencies.
Based on this, our approach is to embed a PTLM in a broader system with an
evolving, symbolic memory of beliefs - a BeliefBank - along with two
mechanisms to improve belief consistency of the overall system.
First a reasoning component -- a weighted
MaxSAT (satisfiability) solver -- reasons about belief dependencies and revises beliefs that
significantly clash with others. Second, a feedback component poses future queries to the model
using known beliefs as context, aiming for more accurate and consistent answers from the PTLM itself.
The BeliefBank represents the overall system's beliefs (a ``mental model'') about
the world, constructed by deliberating over the noisy output of a raw PTLM.

We explore this in a controlled experimental setting where 
both candidate beliefs and constraints between them are provided. Candidate facts are simple
sentences that may be true or false, e.g., "An eagle is a bird" (T), ``An eagle is a mammal'' (F). Constraints 
are between (variabilized) facts, e.g., ``X is a bird $\rightarrow$ X has wings''.
These allow us to both probe and measure improvement in the consistency and accuracy
of a system's beliefs, compared with an original PTLM.

\edited{In contrast to prior work, this system does not rely on fine-tuning the PTLM.
Fine-tuning requires expensive training data, and risks destabilizing the model's performance on other tasks outside the scope of training.
Instead, our system functions without training data, explicitly reasoning about beliefs using an external mechanism, thus allowing both
controllability and interpretability.
Most significantly, we find that improving consistency in this way improves accuracy, while earlier fine-tuning-based approaches report either no accuracy gains \cite{Ribeiro2019AreRR,Minervini2018AdversariallyRN,Li2019ALF} or only slight gains \cite{Asai2020LogicGuidedDA}.}

\eat{
In contrast to prior work this system does not rely on finetuning the PTLM. Finetuning the model requires expensive training and data and would create yet another black box model without guarantee that performance on other tasks remains stable. Our system functions without training data and offers better controlability and interpretability. Most importantly, we show that improving consistency in turn improves accuracy whereas finetuning-based approaches report no \cite{Ribeiro2019AreRR,Minervini2018AdversariallyRN,Li2019ALF} or only slight \cite{Asai2020LogicGuidedDA} accuracy gains.
}

\noindent
We make the following contributions:
\begin{enu}
\item[1.] We show that a PTLM-based system can be given a consistent notion of belief by
  augmenting the PTLM with a global memory -- the BeliefBank -- that can be deliberated over.
  Specifically, we show that two mechanisms -- constraint reasoning and feedback -- 
  improve the overall system's accuracy and consistency over time.
\item[2.] We contribute a targeted dataset to measure a system's consistency against given constraints.
\item[3.] We provide an analysis of the failure modes and directions for future work.
\end{enu}
This work is significant as it is a first step towards PTLM-based architectures that have
a systematic notion of belief, allowing them to construct a more coherent
picture of the world, and improve over time without model retraining.\footnote{Dataset is available at https://allenai.org/data/beliefbank}

\section{Related work \label{related-work}}



The idea that agents should have a belief system dates back to the earliest years
of AI, e.g., \citet{Mccarthy1959ProgramsWC} envisioned representing a system's beliefs as formal
propositions along with a reasoning process to identify what follows.
Multiple subfields of AI have explored ways of representing and updating
beliefs, including in formal logic \cite{Genesereth1987LogicalFO,Moore1983SemanticalCO},
belief revision \cite{de1986assumption,dechter1988belief}, and
uncertainty \cite{Pearl1986FusionPA}. 
  Similarly, work in cognitive science has promoted mental models -- coherent,
constructed representations of the way the world is believed to be -- as
central to understanding and communication \cite{JohnsonLaird1983MentalM,mental-models,Hilton1996MentalMA}.
We draw on these ideas, proposing how they can be layered on top of PTLMs,
here representing beliefs as NL statements rather than formal structures.

Although PTLMs contain extensive world knowledge \cite{Petroni2019LanguageMA, roberts-etal-2020-much},
they can be inconsistent in their answers to probing
questions \cite{ettinger-2020-bert,davison-etal-2019-commonsense,ravichander-etal-2020-systematicity, Elazar2021MeasuringAI,Subramanian2020ObtainingFI}, making their ``world model''
unclear. Although various approaches have improved answer consistency, mainly 
through modified model training, e.g., \cite{Ribeiro2019AreRR,Minervini2018AdversariallyRN,Li2019ALF,Asai2020LogicGuidedDA},
they have not solved the problem. Current PTLMs still behave as a source
of noisy knowledge, rather than projecting a coherent picture
of the world \cite{Kassner2020NegatedAM}.

A close analogy to our task is in knowledge graph (KG) construction.
\citet{Pujara2013KnowledgeGI} define ``knowledge graph identification''
as the task of building a maximally consistent KG given noisy candidate facts
and their extraction confidences, and constraints between them.
They develop a solution using probabilistic soft logic (PSL) \cite{Broecheler2010ProbabilisticSL} as their constraint reasoner.
Our reasoning component follows similar ideas, but applied to the noisy predictions of a PTLM. 
\edited{On the face of it, it is not clear how to plug a constraint solver into a PTLM, given their very different natures.
Our solution introduces a global persistent memory, making this novel combination of technologies possible.
To our knowledge this has not been done before.}

Our work presents a broader system architecture
in which a PTLM is embedded, along with a dynamic, persistent memory.
While there are prior neural architectures that include an associated memory,
e.g., \cite{Henaff2017TrackingTW,Sukhbaatar2015EndToEndMN,Graves2016HybridCU},
these components typically play the role of a short-term working memory to help computation.
In contrast, our BeliefBank is a persistent, long-term memory, 
and we treat the PTLM as a component within a larger system.
Our work also differs from retrieval-augmented architectures, 
e.g., RAG \cite{Lewis2020RetrievalAugmentedGF}, REALM \cite{Guu2020REALMRL},
that augment model input with external retrievals. Rather, 
our memory is {\it reflective}, built from model {\it outputs} and reasoned over.

Our feedback mechanism uses old answers to help answer new questions.
This builds on prior work such as Self-Talk \cite{selftalk},
where a model asks itself related questions to help with new answers, 
\edited{and can be seen as a form of dynamic prompt engineering \cite{Liu2021WhatMG}}.
In our case, feedback is selected from a global BeliefBank, rather
than generated with templated subqueries, potentially allowing more 
control over feedback selection.

Finally, our system performs a kind of continual learning \cite{Parisi2019ContinualLL,Carlson2010TowardAA}.
While recent work in this area has focused on dynamic update of model parameters, e.g., \cite{Ebrahimi2020RememberingFT},
our work leaves the model fixed, and seeks improvement in the broader system in which the
model is embedded, exploring an alternative and potentially more interpretable architecture
towards this goal.

\section{Task}



Our goal is to ascribe a clearer notion of ``belief'' to a system that includes a model $M$,
by improving, over time,  the consistency and accuracy of
its answers (compared with $M$)
to a stream of questions. We measure this
with true/false probing, where we are also given a set of
constraints between answers:
\begin{myquote}
{\bf Given:}
\begin{ite}
\item a stream of {\bf sentences} $Q$, interpreted as
questions to the model 
\item a set of {\bf constraints} $C(S)$ between (the truth values of) sentences in $Q$, each annotated with a weight $w_i$ (A penalty $w_i$ is applied if $c_i \in C(S)$ is violated)
\item a {\bf model} $M$ that takes as input a  question
$q \in Q$ and optionally an (NL) context $X$ (consisting of
answers to previously posed questions), and predicts a True/False answer $A$ with confidence score $F$
\end{ite}
{\bf Accumulate:}
\begin{ite}
\item
the True/False labels for $Q$ predicted by $M$,
optionally corrected by the constraint solver,
so as to maximally improve accuracy (with respect to gold labels) and
   consistency (minimize total penalties of constraint violations) 
\end{ite}
\end{myquote}

\section{Approach}

Our approach adds a memory layer, called the {\bf BeliefBank}, on top of the model to globally 
track beliefs. Two mechanisms are then used to manage the BeliefBank beliefs, namely
constraint reasoning and feedback, as we now describe.

\subsection{Definitions \label{definitions}}

\noindent Let
\vspace{-2mm}
\begin{ite}
  \item a {\bf belief} $b_i$ be a triple ($s_i$,$l_i$,$w_i$), where 
     \begin{ite}
     \item $s_i$ is a sentence 
     $\in S$ 
      \item label $l_i \in$ \{T,F\} denotes the system's True/False belief about the truth of $s_i$\footnote{
            Strictly, the label F denotes the belief that the negation of  $s_i$ is true,
      e.g., (``a poodle is a bird'',F,...) denotes the belief ``a poodle is not a bird''.}
     \item weight $w_i$ is a number $\in [0,1]$ representing the system's strength of that belief
    \end{ite}
 For example:
 \begin{quote}
{\it      ("a poodle is a dog", T, 0.9) }
 \end{quote} 
 denotes the belief (strength 0.9) that "a poodle is a dog" is a true statement (T).
  \item a {\bf BeliefBank} $B(S)$ = a set of beliefs over sentences $S$ = $s_1,...,s_n$

\item a {\bf constraint} $c_i$ = a 5-tuple of the form ($s_i.l_i \rightarrow s_j.l_j,  w_i$)
where 
\begin{ite}
\item $s_i,s_j$ are sentences $\in S$, 
\item $l_i,l_j \in$ \{T,F\}.
If $s_i$ has truth value $l_i$, denoted $s_i.l_i$, then $s_j$ is expected to have truth value $l_j$, denoted $s_j.l_j$.
\item $w_i$ denotes the strength of that expectation (a penalty $w_i$ is applied if violated). 
\end{ite}
For convenience, a shared variable X can be used in $s_i,s_j$, allowing a set of grounded
constraints to be expressed in one statement, e.g.,
\begin{myquote}
{\it (``X is a dog''}.T $\rightarrow$ {\it ``X has a tail''}.T, 0.8) 
\end{myquote}
expresses that if something is a dog, then it should (T) have a tail, with a penalty of 0.8 applied if
it does not. Mutual exclusivity is expressed using two rules, e.g., that fish and birds are mutually exclusive
is expressed:
\begin{myquote}
{\it (``X is a bird''}.T $\rightarrow$ {\it ``X is a fish''}.F, 1.0)  \\
{\it (``X is a fish''}.T $\rightarrow$ {\it ``X is a bird''}.F, 1.0) 
\end{myquote}
where ``F'' indicates the conclusion should be {\it false} if the condition here is true (T).
\item a {\bf constraint graph} $C(S)$ = a set of constraints $c_i$ over sentences $S$ 
\end{ite}
Given a set of beliefs $B(S)$ about $S$ and a set of constraints $C(S)$, we measure
{\bf consistency} using (the complement of) \citet{Li2019ALF}'s conditional 
constraint violation ($\tau$) metric,
namely the fraction of constraints whose {\it condition} $s_i.l_i$ is believed,
but whose {\it conclusion} (that $s_j$ has truth value $l_j$) is not. 
In other words, over all constraints $c_i \in C(S)$, inconsistency $\tau$ is

\noindent
\hspace*{2mm} $\tau = | \{ ~c_i~ |~ \neg (s_i.l_i \rightarrow s_j.l_j)~ \} | ~~ / ~~ | \{~ c_i ~|~ s_i.l_i ~\} |$ 

\noindent
i.e., the size of the set of {\it violated} constraints ($s_i.l_i \rightarrow s_j.s_j$ is false)
divided by the size of the set of {\it applicable} constraints. 
We then define:
\begin{myquote2}
\begin{center}
consistency = 1 - $\tau$
\end{center}
\end{myquote2}

\subsection{Methods}

We consider our system in a dynamic setting, where it receives a stream of questions
and gradually builds up a BeliefBank of answers (including revising earlier
answers). We evaluate two methods for improving the BeliefBank's accuracy and consistency
over time:
\begin{des}
\item[{\bf Constraint solving:}] Given a model $M$'s raw answers (with confidences), a constraint
solver seeks to reduce constraint violations by potentially flipping answers that maximally clash
with other answers. 
\item[{\bf Feedback:}]
Given a new question $q$, selected beliefs in the BeliefBank 
are provided as context to $M$ to help it answer $q$ correctly.
\end{des}
\vspace{1mm}
Figure~\ref{architecture} shows these components.

\subsubsection{Constraint Solving \label{constraint-solving}}

Given a set of beliefs and constraints, the constraint solver has two competing objectives: (a) flip 
beliefs so as to minimize constraint violations (b) don't flip beliefs, so as to preserve
the model's raw answers, i.e., minimize conflict between the model and BeliefBank. 
To implement this tradeoff, the model's answers are themselves treated as just
another constraint, e.g., the answer that "a poodle is a dog" is true (confidence 0.9) is treated as a constraint
"a poodle is a dog", with penalty 0.9 if it is violated. 
To balance the two objectives (a) and (b), the model confidences are scaled by a learned hyper-parameter $\lambda$,
trained on a calibration part of our dataset, disjoint from the data then used in experiments (Section~\ref{dataset}).

To implement constraint solving, we translate the task into a {\it weighted MaxSAT} (satisfiability) problem $P$,
for which efficient algorithms with guarantees exist. Each belief becomes a weighted assertion in $P$, e.g.,
the belief ("a poodle is a dog", T, 0.9) is expressed in SAT syntax:
\begin{myquote2}
{\it 0.9 } 
{\it "a poodle is a dog"}\footnote{In practice, strings are replaced with numeric identifiers in SAT syntax, but for clarity we leave them as strings here.}
\end{myquote2}
while the constraint ("a poodle is a dog".T $\rightarrow$ "a poodle has a tail".T, 0.8) is expressed:
\begin{myquote2}
{\it 0.8 "a poodle has a tail"   ~{\bf -}"a poodle is a dog"}
\end{myquote2}
(literally: {\it "a poodle has a tail"} OR NOT ("{\bf -}") {\it "a poodle is a dog"}). 
We then apply the solver Z3 \cite{z3} to $P$, which outputs a set of truth assignments for all individual 
sentences in $P$ so as to minimize the weighted sum of violations. If the truth  of any sentence
has changed, the BeliefBank is correspondingly updated. 


\subsubsection{Feedback \label{feedback}}

Feedback involves asking the model a question, but with the benefit
of knowing answers to prior, related questions. To use these answers in the query,
selected beliefs are added to the query context before asking the model.
(Note that the selected beliefs are not guaranteed to be correct, of course).
Our conjecture is that if the model is explicitly reminded of relevant beliefs when answering a new question,
it will answer the question more accurately and consistently.
For example, in Figure~\ref{architecture}, when asked "Do swallows have gills?", our model $M$ incorrectly answers "yes".
But if reminded that swallows are not fish, by asking: "CONTEXT Swallows are not fish. QUERY Do swallows have gills?" 
the model now correctly answers "no". 

We evaluate two policies for choosing which beliefs to feed back to $M$ when asking question $q$
about entity $e$:
\begin{enu}
\item[1.] {\bf on topic} beliefs, namely current beliefs about entity $e$, randomly selected from the BeliefBank
\item[2.] most {\bf relevant} on topic beliefs (i.e., again about $e$), using the constraint graph to identify relevance.
As the constraint graph captures potential clashes that the answer to  $q$ could cause, we use the graph to identify beliefs that would be most affected by that answer. 
For example, if the current query is: "Is a poodle an animal?", the constraint graph identifies potential clashes that
would occur if the model
answered "yes", and also clashes if it answered "no". Here, if the model answered "no", the resulting belief ("a poodle is {\it not} an animal") would strongly clash with other beliefs "A poodle is a dog." and "A poodle is a mammal.", 
so these two are strong candidates for the context. We select the three strongest clashing beliefs found in this way, considering both "yes" and "no" answers to $q$. If no relevant fact is present, we use a randomly selected topic belief instead.

\end{enu}
In both cases, three beliefs are selected, this number was empirically found to be most effective.

\section{Dataset \label{dataset}}


We create a dataset\footnote{Dataset is available at https://allenai.org/data/beliefbank}
to test our approach in a controlled way, allowing us to perform systematic experiments to evaluate behavior.
The dataset contains two parts, {\it constraints} and {\it facts}, defined over simple sentences
such as ``a swallow is a bird.''

\subsection{Constraints \label{dataset-constraints}}

The dataset contains two kinds of constraints: 
\begin{des}
\item[{\bf positive implications:}] conclusion truth value $\textrm{l}_j$ = T (true), e.g.,
\begin{myquote}
{\it ``X is a dog.T $\rightarrow$ ``X has a tail.''.T}
\end{myquote}
\item[{\bf mutual exclusivities:}] expressed as a pair of constraints with $\textrm{l}_j$ = F (false), e.g.,
\begin{myquote}
{\it ``X is a dog".T $\rightarrow$ "X is a bird.''.F} \\
{\it ``X is a bird".T $\rightarrow$ "X is a dog.''.F}
\end{myquote}
expresses that an entity cannot be both a dog and a bird at the same time.

\end{des}
Positive implications were manually gathered from ConceptNet \cite{Speer2017ConceptNet5A}.
First, we identified 121 general concepts of interest, e.g., ``mammal'', then
converted selected triples about them to constraints (Details of the selection process are in Appendix~\ref{conceptnet}).
For example, the ConceptNet triple (dog,HasA,tail) becomes the
constraint "X is a dog" $\rightarrow$ "X has a tail".
We also add weaker, disjunctive constraints in the backward direction, e.g., 
"X has a tail" $\rightarrow$ "X is a dog" {\it OR} "X is a cat" {\it OR} ....  
for all entities with tails.
Mutual exclusivities were gathered from the ``isa'' taxonomies in ConceptNet and WordNet \cite{wordnet},
using the approximation that
siblings in the noun hierarchy are mutually exclusive.
Thus, for any pair of siblings,
we add a mutual exclusivity constraint (using two constraint rules). 

We collected 2612 constraints in this fashion (1836 forward implications, 2*388 bidirectional mutual
exclusivities).

\subsection{Constraint Weights}
\label{section:penalties}
Constraint weights need to be set appropriately
to mix well with the model's confidences inside the weighted SAT solver. We use a development set of 1072 facts about seven entities to set one constraint weight for the forward direction of the implications and the mutual exclusivity rules and a second one for the backward direction of the implications. To do this we perform a grid search over these parameters, finding the values that result in
the highest F1 (accuracy) after running the constraint solver over the raw model's beliefs about
these facts.

\subsection{Facts}

We also collect a set of truth-labeled facts about different entities, relevant to the constraints.
To do this, we select a new entity, e.g., "poodle", that is a member of one of our general concepts,
e.g., "dog", then instantiate the constraint graph with that
entity (i.e., set X = "poodle"). We then identify the leaf (source) nodes of that graph, just considering
forward implication rules, i.e., finding facts not implied by other facts in the graph, and manually
annotate their True/False labels. We then use the implications and mutual exclusivities to infer
other True/False labels for other sentences, i.e., we propagate the annotated labels through the graph.
This provides ``silver'' labels for sentences reachable in this way (a subset of all the sentences 
in the graph) -- silver because the implications are soft, hence not guaranteed to hold for all entities.

We repeat this for 85 entities (animals and plants), resulting
in a final dataset containing 12,525 ``silver'' facts (sentences + True/False labels). Note that this data is purely for evaluation. There is no training phase or training data. The system does not have access to any labeled data besides the constraint rules.

\section{Model}
The fixed model $M$ that we use for our experiments is Macaw \cite{macaw}, a state-of-the-art T5 QA model fine-tuned on $\approx$400k
QA pairs. 
To query the model, we pose the query (optionally with a textual context), and let the model choose between the two answer options "yes" and "no".
The model also outputs an answer confidence, used as the belief weight. 

We use the T5-large version of this model. Note that we do not retrain the model for
this work; rather, it is used as a black-box QA module in the broader system (Figure~\ref{architecture}).
Other models could equally have been used.

\section{Experiments}

\begin{figure}
\centering
     \includegraphics[width=1\columnwidth]{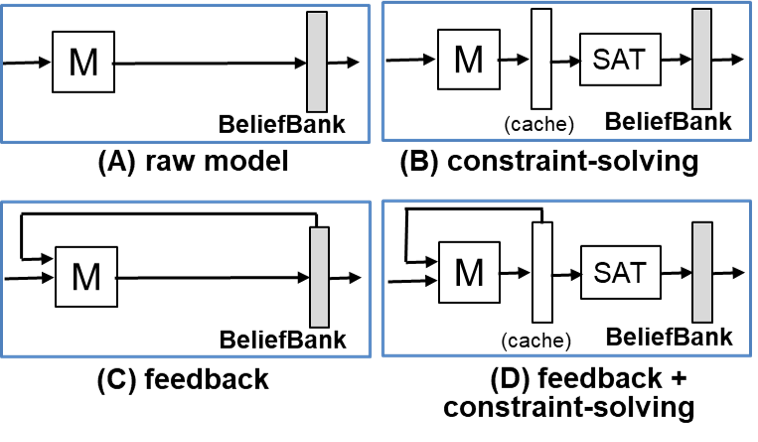}	   
    \vspace{-6mm} \\
     \caption{The four configurations we evaluate. 
       In (B), the contraint-solver (SAT solver) is run over all model M answers so far. 
       In (C), current beliefs are fed back as context for new questions.
       (D) combines the two.
       \label{configurations} 
       }
       \vspace{-4mm}
\end{figure}

We evaluate our system in a dynamic setting in which it receives a stream of questions,
building up and revising a BeliefBank.
To simplify the evaluation, we consider questions to arrive in {\it batches},
and evaluate the BeliefBank after each batch, measuring accuracy (F1)\footnote{
We measure accuracy with F1 (on the True class) rather than \% correct because
the True/False distribution in our dataset is unbalanced, with significantly fewer True than False
answers. F1 avoids scores being dominated by negative answers.}
and consistency (1-$\tau$, Section~\ref{definitions}) of the BeliefBank so far,
comparing with the gold labels. 
We evaluate four configurations:
\begin{des}
\item[{\bf Raw model:}] The BeliefBank simply records the raw model's answers \footnote{To the best of our knowledge there are no other baseline models to compare to as consistency based Q\&A does not go beyond paraphrases and relies on finetuning \cite{Elazar2021MeasuringAI}.}
\item[{\bf Constraint-Solving:}] After each batch, the constraint solver is run
    over all the (raw) model answers so far, and the BeliefBank updated accordingly.
\item[{\bf Feedback:}] Questions in batch $n$ are posed to the model using a
    context selected from the beliefs already in the BeliefBank (batches 1 to $n-1$).
    We evaluate two selection strategies:
    \begin{des}
    \item[{\bf Feedback (on-topic):}] Random beliefs about the entity $e$ being queried about
    \item[{\bf Feedback (relevant):}] On-topic beliefs (i.e., again about $e$) that are most relevant to the query, as defined in Section~\ref{feedback}
    \end{des}
\item[{\bf Feedback + Constraint-Solving:}] A combination of the two.    
\end{des}
\noindent
These configurations are illustrated in Figure~\ref{configurations}.

\begin{figure*}[t]

  \includegraphics[width=0.49\textwidth]{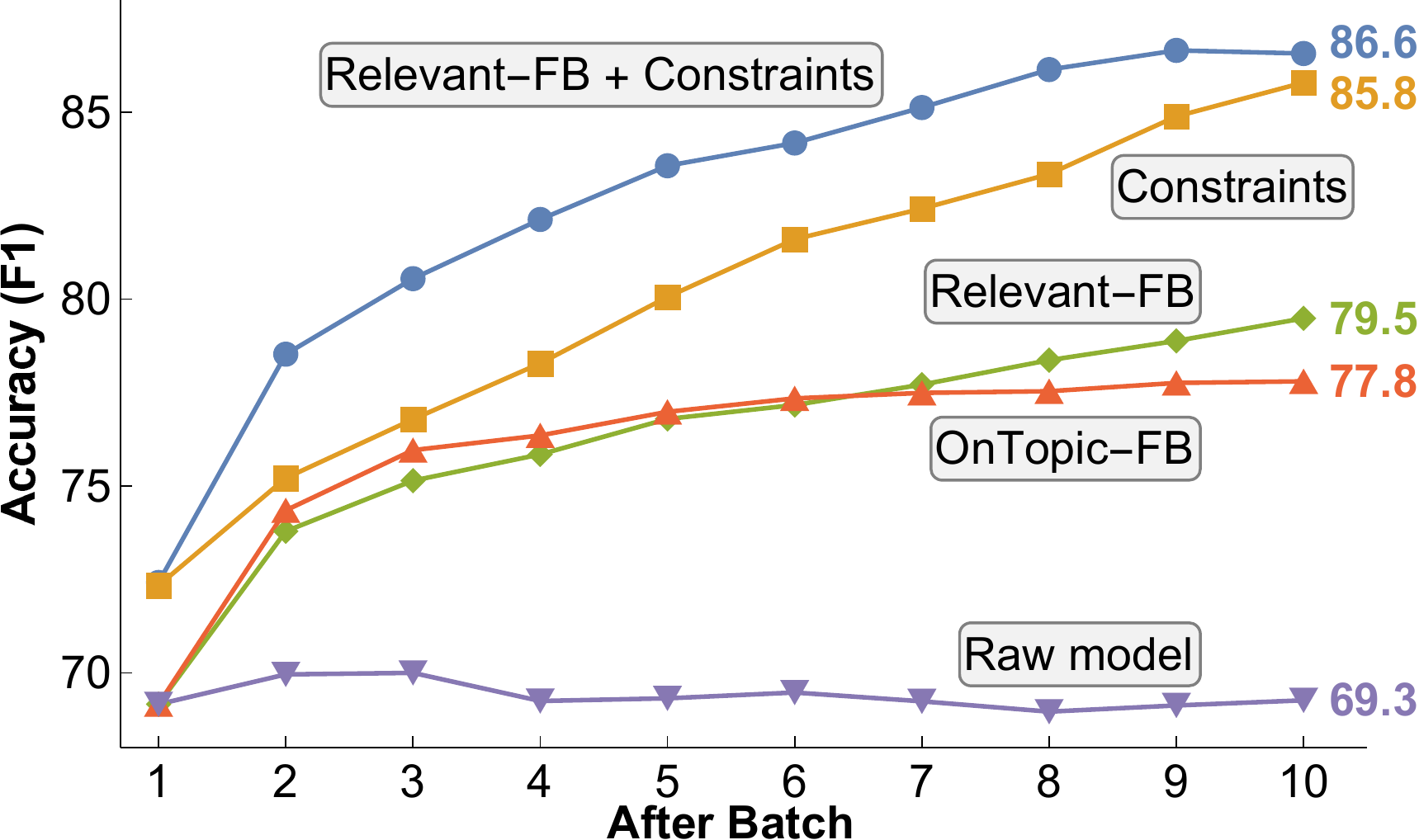}
  \hspace{2mm}
    \includegraphics[width=0.49\textwidth]{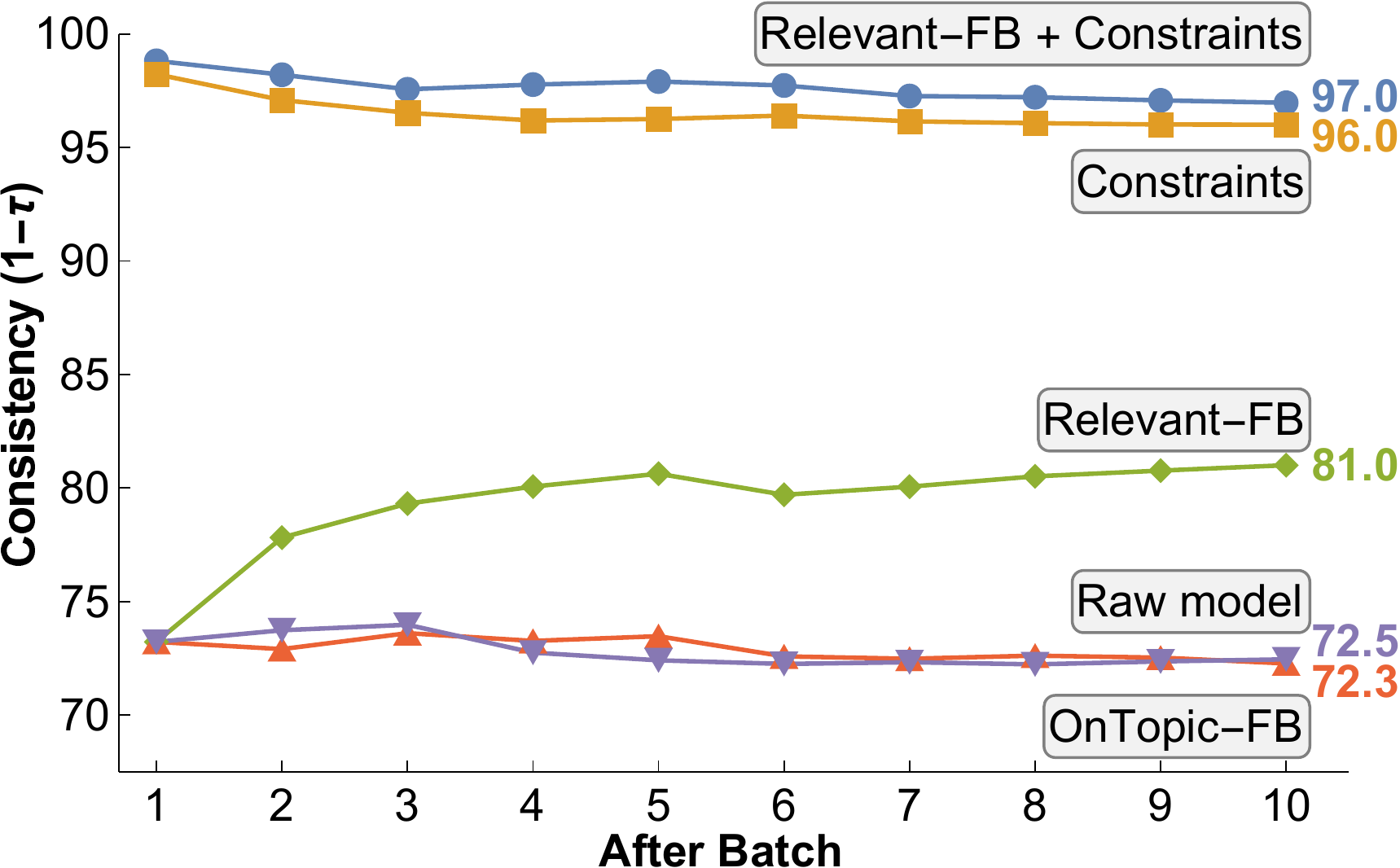}
\vspace*{1mm}
 {\fontsize{10}{8}\selectfont
\hspace*{1.5cm} {\bf OnTopic-FB} = using (randomly selected) {\bf on-topic} feedback from old answers for new queries. \vspace*{-1mm} \\
\hspace*{1.5cm} {\bf Relevant-FB} = using most {\bf relevant} on-topic feedback for new queries. \vspace*{-1mm} \\
\hspace*{1.5cm} {\bf Constraints} = running the constraint-solver after each batch. \vspace*{-1mm} }
\caption{Accuracy (left) and consistency (right) of the growing BeliefBank, as the system answers incrementally more questions (each batch = 10\% of the queries).
  Relevant feedback, constraint-solving, and both, all help improve both F1 and Consistency.  \label{results}}

\vspace{-2mm}
\end{figure*}

\subsection{Results}

The results are shown in Figure~\ref{results}, showing the changing accuracy and consistency of the growing
BeliefBank with time, for different configurations. Each time-step (batch) represents another 10\% of the test questions
being posed to the system. (The same data is presented in tabular form in Appendix~\ref{tables}).
Several conclusions can be drawn:


\mybullet{} Use of feedback, constraint-checking, or both, all result in a {\bf continually improving accuracy over time}.
    This is a significant result, showing a larger system can continually improve even if its internal PTLM component is fixed.
    (The raw accuracy of the PTLM itself is necessarily constant)\eat{, as questions are posed independently and in random order \nk{Why would the order and dependency have an affect on the raw answers?})}.

\mybullet{} {\bf Use of the constraint-solver results in very high (\textasciitilde 95\%) consistency,} indicating
that it is doing its job well, and {\bf also improving accuracy} substantially (+17\% over the raw model). The constraint-solver has a global view of the BeliefBank, and thus can balance all beliefs seen so far with the provided constraints to make a decision.

  \mybullet{} {\bf Relevant feedback results in significant consistency gains} compared with just on-topic feedback.
  As relevant beliefs are exactly those that may clash with the answer to the current question (Section~\ref{feedback}),
  this encourages the model to answer consistently with those beliefs, promoting consistency. Of course, this
  could hurt accuracy if those relevant beliefs were wrong, amplifying the errors. In fact, 
  the overall accuracy remains about the same as with on-topic feedback, and significantly better than the
  model's raw answers.
  
\eat{
\mybullet{} {\bf On-topic and relevant feedback result in similar accuracy gains} of +7\% F1 over the raw model. Accuracy and consistency gains are lower than for the constraint solver as feedback is only able to exploit part of the belief bank,
and (unlike the constraint solver) cannot modify earlier answers.
  It is perhaps surprising that our relevance-selection method does not provide more gains over
  feedback that is simply on-topic (but otherwise selected randomly). We do see that the resulting
  {\bf consistency with relevant feedback is significantly higher than with on-topic feedback},
  possibly because \red{Why?} \nk{relevant sees belief in the context that could potentially clash with the current query}.
  }
  
\mybullet{} The {\bf greatest gains are for feedback and constraint-solving combined}, resulting in +18\% F1 (absolute)
over the raw model accuracy. This suggests that feedback and constraints can work together in a positive way.

\eat{This is currently more of a problem using feedback: Note that accuracy improvement
is not guaranteed, as the constraint solver could (and sometimes does) flip correct beliefs to
incorrect, so as to improve overall consistency. We discuss this in more detail shortly.}


\eat{
\mybullet{} To see whether simply {\it any} feedback helped the model, we repeated
the feedback experiments but using randomly selected facts drawn from the predictions of {\it any} entity in the dataset
({\bf random} \nk{should we call it off-topic?} feedback). This feedback did not help,
the system performing similarly to the raw model (point (B) in Figure~\ref{results}). 
indicating that {\bf the on-topic nature of the feedback is important}.
}
  \mybullet{} As a sanity check we also tried using random beliefs about {\it other} entities (off-topic)
  as feedback, but (as expected) the results did not significantly differ from no feedback (raw model),
  ending at $\approx$72\% F1 and $\approx$74\% consistency after batch 10. %

We also evaluated a {\bf non-incremental}, ``omniscient'' version of
the BeliefBank: Given the raw model answers to {\it all} questions, re-ask
every question using feedback selected (using relevance) from all the other answers.
The resulting accuracy was 74.5\%, 
substantially lower than for the on-topic incremental approaches.
Interestingly, this approach's built-in advantage (that every question has
access to answers for {\it all} other questions) does not outweigh the
built-in disadvantage (that those are the raw, rather than
incrementally corrected, answers). 
This is a significant result demonstrating that
{\bf the positive feedback loop of the incremental approaches can be advantageous},
where feedback feeds more accurate beliefs into the BeliefBank,
improving future feedback, etc.


\eat{We also evaluated a non-incremental, ``omniscient'' version of
the BeliefBank: Given the raw model answers to {\it all} questions, re-ask
every question but with context feedback selected from all the other answers
(using the most effective selection method, namely relevant feedback).
The resulting performance is shown as point (A) in Figure~\ref{results}.
Interestingly, although this approach has a built-in advantage (as
every question has access to answers to all other questions), the
result was {\bf less consistent than using relevance feedback incrementally}
(which scored +5\% higher absolute), and with lower accuracy than using
the constraint-solver. This indicates that incrementally building up
a belief bank can outperform an all-at-once approach, at least for some
configurations.
}

\subsection{Failure Analysis}

We now provide some examples of good and bad flips to
better understand the behavior of the model.

First, as an {\bf illustration of desired behavior}, the raw model incorrectly believes that a pine is both a plant (correct) and a
vertebrate (incorrect), when queried. However, this violates a mutual exclusivity rule, so the constraint-solver
considers flipping one of these. Flipping ``pine is a plant'' from T to F would result in 
numerous other violations, e.g., ``pine is a tree'' (which the model also believes)
would be violated. As a result, it prefers to (correctly) disbelieve ``pine is a vertebrate'', improving
both  accuracy and consistency. 

From an analysis of the data, we see that the
{\bf majority of the raw model errors are false positives} -- the raw model
generally answers (almost) all the positive facts correctly (recall is $\approx$98\%),
but mistakenly thinks many negative facts are also true (precision is $\approx$54\%).
These false positives can be rather unusual facts, e.g., ``A poodle is a bathroom.'' (model's answer: True).  
It is unsurprising that the model knows most of the positive facts, as they are simple statements about 
common entities (``eagles can fly''), likely seen in pre-training. However, the fact that the model 
makes (what a person would view as) catastrophic errors when asked more unusual
questions, e.g., believing that ``a poodle is plant'', 
reveals that the PTLM's grasp of the world is still incomplete and problematic. The constraint mechanism
proposed here essentially asks the model to think about its answers and their consequences,
so that it can spot problems that the PTLM alone does not see, and repair them. 

The constraint reasoner can also make mistakes, flipping things the wrong way
so as to improve consistency, at the expense of accuracy. For example, the raw model
correctly believes that  ``a rat is not a cat''. However, the constraint solver
then (incorrectly) flips this to ``a rat is a cat'', because multiple constraints
weakly suggest rats are cats given other beliefs (''rats catch mice'', ``rats have tails'',...),
which together add up, causing the (incorrect) flip, including overwhelming the
strong (but not infinitely strong) constraint that ``a rat is not a feline.'' This
illustrates that the {\bf constraint mechanism is sensitive
  to the number of and weights on constraints},
even with automatic hyperparameter tuning (Section~\ref{section:penalties}).

Similarly, {\bf the feedback mechanism is sensitive to question order},
especially if the model's early answers are wrong, as the feedback mechanism
causes the model to pay extra (sometime disproportionate) attention to earlier context \cite{Kassner2020NegatedAM}. For example,
the bad context ``A poodle is not a mammal'' (from an earlier bad answer) undesirably causes
the model to change its answer for "A poodle is a dog" from true (raw model) to false.

Finally, we can only speculate why feedback improves
results, in particular, since  the feedback consists of facts that came from the model itself (i.e., that it already knows).
One explanation is that {\bf feedback may help the model focus attention} on important facts,
e.g., reminding the model that ``a swallow is not a fish'' should help it realize
that ``a swallow has gills'' is False (Figure~\ref{architecture}). In addition,
feedback may possibly help resolve some ambiguities, e.g., the feedback
``a swallow has wings'' helps identify the bird sense of ``swallow''.
Similar advantageous use of feedback was observed in the SelfTalk
experiments \cite{selftalk}.

In future work, the feedback mechanism can be improved further by training it to respond more systematically to feedback (similar to \cite{ruletaker}) and to better balance implicit and explicit knowledge  \cite{DBLP:conf/nips/TalmorTCGB20}, ideally incorporating different levels of confidence.

\section{Future Work}


\subsection{Human in the Loop}

Although our system is autonomous, its incremental setting 
combined with the explicit representation of beliefs makes it amenable to a human in the loop.
In this setting, a human might spot an egregious bad belief in the BeliefBank, and forcibly
correct it. Then, ideally, this strong positive datapoint would {\it also} improve the model's
accuracy on other beliefs, both in the BeliefBank and for future questions. As a brief test
of this, we allowed a human to correct all bad beliefs (average 6) in
the BeliefBank after just the first batch (10\%) of questions, and then continued
as before to completion, using the constraint-solving approach. We find that these limited interventions increased 
both the final F1 and Consistency each by 2\% (absolute) {\it on top of} the gains produced by the corrected beliefs themselves.
Although preliminary, this suggests that our architecture may have value in an interactive
``machine teaching'' setting, where the user is supervising and correcting the system,
and it continually improves as a result \cite{Zhu2015MachineTA}.

\subsection{Towards Deployment}

Although our work has been in a constrained setting (targeted set of relations, entities and constraints), there is a clear development path
to deployment in real QA systems to reduce the kind of irrational behavior we have described,
such as in this (real) transcript:
\begin{myquote}{\it 
(1) Is oxygen colorless? yes \\
(2) What color is oxygen? blue \\
(3) What gas do plants produce? oxygen \\
(4) What color is the gas plants produce? green}
\end{myquote}
The basic components of our architecture provide a framework to help avoid such irrationality.
First, (declarative versions of) questions and model answers would be persistently stored in a BeliefBank.
Second, on-topic feedback could be selected to help answer new questions using
information retrieval over the BeliefBank. Third, given a source of constraints, e.g.,
a general rule of taxonomic inheritance,\footnote{I.e., that the properties of an entity type usually apply to all its subtypes also.}
constraint solving could be applied to spot and reduce clashes. This would
require a mechanism to identify when a belief satisfies a constraint's
condition or conclusion, e.g., a state-of-the-art textual entailment engine
such as CA-MTL \cite{Pilault2020ConditionallyAM}. A variant of our system could also work without the distinction of model beliefs and constraints: Instead of providing constraints externally we could treat them as beliefs, e.g., query the model for mutual exclusivities "Can an entity be an animal and a plant?" or implications: "Do dogs have tails?" directly. 
\edited{This would run the risk of adding extra noise, but would eliminate the manual effort involved in generating the constraint set, and therefore improve scalability.
Together, such developments would
pave the way to real-world QA systems that are more consistent and
improve over time, rather than remain static.}

\subsection{The Broader Research Agenda}

This work only touches on a broader research agenda, namely
how to expand work on PTLMs
to encompass the cognitive skills of world modeling and deliberative reasoning
(``thinking, fast and slow'' \cite{Kahneman2011ThinkingFA}).
In this broader agenda, intelligence is not just about opaque question-answering, but
also about constructing mental models that describe how (some aspect of) the world works \cite{mental-models}. 
Although mental models are abstractions (hence are approximate), they add a powerful,
systematic component to understanding that should expand its capabilities.

The BeliefBank can be seen as a simple illustration of this broader agenda.
A wider pursuit would include a richer notion of a model, perhaps with
more structure to model elements than just sentences; more sophisticated means of
model construction than just accumulating and resolving answers; and the generation of
explanations to convey the deliberative component's behavior, and ultimately
interact with a user. Such mechanisms may be symbolic or neural in
nature, e.g., \cite{DBLP:conf/nips/TalmorTCGB20}.
Although these issues are beyond the scope of this paper, our work
points to this interesting, larger goal for PTLM research, as
well as offering a specific mechanism for belief consistency. 

\section{Conclusion}

PTLMs can be inconsistent in their answers to probing questions, and
can still give (what to a person appear as) naively wrong answers. This work is a first step towards alleviating these
problems.  
By embedding a PTLM within a larger system with a persistent, global memory -- the BeliefBank --, a constraint-solver and feedback mechanism, we have shown
that the overall system's behavior is more coherent, both
in terms of consistency and accuracy.
The additional memory layer can loosely be seen as the system's ``mental model'',
a representation constructed from the PTLM's raw answers.

Our experiments were conducted in a restricted (small set of relations, entities and constraints), controlled setting,
and further development is needed to scale to larger and more
complex tasks. Nevertheless, the work here is significant as it
is a first step towards PTLM-based architectures with 
a globally consistent notion of belief,
allowing them to construct a more coherent picture of the world,
and continually improve with time.

\section*{Acknowledgements}

This work has been funded by the Allen Institute for AI and the European Research Council (\#740516) and by the German Federal Ministry of Education
and Research (BMBF) under Grant No. 01IS18036A. The authors of this
work take full responsibility for its content.


\bibliography{custom,anthology}
\bibliographystyle{acl_natbib}

\clearpage

\appendix

\twocolumn[{\centering {\Large {\bf Appendix: BeliefBank: Adding Memory to a Pre-Trained Language \vspace{1mm} \\ Model for a Systematic Notion of Belief \vspace{5mm}}}}]

\begin{table*}[t!]
\centering
{\small
\begin{tabular}{lcccccccccc} \hline
Accuracy (F1) after batch $\rightarrow$ & 1 & 2 & 3 & 4 & 
5 & 6 & 7 & 8 & 9 & 10\\ \hline           
Raw model & 69.2 & 70.0 & 70.0 & 69.3 & 69.3 & 69.5 & 69.2 & 69.0 & 69.1 & 69.3\\
OnTopic-FB & 69.2 & 74.3 & 76.0 & 76.4 & 77.0 & 77.3 & 77.5 & 77.5 & 77.8 & 77.8\\
Relevant-FB & 69.2 & 73.8 & 75.1 & 75.8 & 76.8 & 77.2 & 77.7 & 78.4 & 78.9 & 79.5\\
Constraints & 72.3 & 75.2 & 76.8 & 78.3 & 80.1 & 81.6 & 82.4 & 83.3 & 84.9 & 85.8\\
Relevant-FB + Constraints & 72.4 & 78.5 & 80.5 & 82.1 & 83.6 & 84.2 & 85.1 & 86.1 & 86.7 & 86.6\\\hline
\end{tabular}
}
\caption{Experimental results for accuracy (F1), as plotted in Figure~\ref{results}, here shown in tabular form.}
\label{table-f1}
\vspace{-3cm}
\end{table*}

\begin{table*}[t!]
\centering
{\small
\begin{tabular}{lcccccccccc} \hline
Consistency ($1-\tau$) after batch $\rightarrow$ & 1 & 2 & 3 & 4 & 
5 & 6 & 7 & 8 & 9 & 10\\ \hline           
Raw model & 73.2 & 73.7 & 74.0 & 72.8 & 72.4 & 72.3 & 72.3 & 72.2 & 72.4 & 72.5\\
OnTopic-FB & 73.2 & 72.9 & 73.6 & 73.3 & 73.5 & 72.6 & 72.5 & 72.6 & 72.5 & 72.3\\
Relevant-FB & 73.2 & 77.8 & 79.3 & 80.1 & 80.6 & 79.7 & 80.1 & 80.5 & 80.8 & 81.0\\
Constraints & 98.2 & 97.1 & 96.5 & 96.2 & 96.3 & 96.4 & 96.1 & 96.1 & 96.0 & 96.0\\
Relevant-FB + Constraints & 98.8 & 98.2 & 97.6 & 97.8 & 97.9 & 97.7 & 97.3 & 97.2 & 97.1 & 97.0\\\hline
\end{tabular}
}
\caption{Experimental results for consistency ($1-\tau$), as plotted in Figure~\ref{results}, here shown in tabular form.}
\label{table-cons}
\vspace{-3cm}
\end{table*}

\section{Selecting Constraint Rules from ConceptNet \label{conceptnet}}

As described in Section~\ref{dataset-constraints}, positive implication (constraint) rules were
manually gathered from ConceptNet \cite{Speer2017ConceptNet5A}. 
First, we identified 121 general concepts of interest, e.g., ``mammal'', choosing concepts
with high occurrence ($>$ 100 times) in ConceptNet, avoiding significantly ambiguous terms (e.g., ``bat''), and filtering out plurals and 
obscure concepts.
For these entities, we then collected all ConceptNet facts involving 6 relations: IsA, HasA, MadeOf, PartOf,
HasProperty, and CapableOf, and
re-expressed them as constraints. For example, the ConceptNet triple (dog, HasA, tail) gives rise to the constraint
"X is a dog" $\rightarrow$ "X has a tail." 
(Triples are converted into English sentences using simple templates).
We then manually filter theses constraints for factual correctness.
We also add weaker, disjunctive constraints in the backwards direction, e.g., 
"X has a tail" $\rightarrow$ "X is a dog" {\it OR} "X is a cat" {\it OR} ....  
for all entities with tails.
(These backwards rules discourage the trivial solution that everything is false.)
Finally, two hyperparameters for weights on forward and backwards rules are
set by automatic calibration (Section~\ref{section:penalties}).

\section{Experimental results in table form \label{tables}}

Tables~\ref{table-f1} and \ref{table-cons} contain the numerical data for the experimental results plotted in Figure~\ref{results}.

\flushend

\end{document}